\documentclass[runningheads]{llncs}
\titlerunning{ADAPT for Faithful MLLMs}
 
\usepackage{eccv}

\usepackage{eccvabbrv}
\usepackage[ruled,vlined]{algorithm2e}

\usepackage{graphicx}
\usepackage{booktabs}
\usepackage{multirow}
\usepackage[accsupp]{axessibility}  

\usepackage{wrapfig}

\usepackage{hyperref}

\usepackage{orcidlink}

\usepackage[table]{xcolor}

\usepackage{adjustbox}
\usepackage{subcaption}
\usepackage{booktabs,multirow}
\usepackage[table]{xcolor}

\begin{document}


\title{ADAPT: Attention Dynamics Alignment with Preference Tuning for Faithful MLLMs}


\author{
Zhiyuan Yao\inst{1,3} \and
Zheren Fu\inst{1} \and
Zhixiao Zheng\inst{1} \and \\
Jiajun Li\inst{2} \and
Yi Tu\inst{2} \and
Zhendong Mao\inst{1,3}\thanks{Corresponding author.}
}
\authorrunning{Z. Yao et al.}

\institute{
University of Science and Technology of China, Hefei, China
\and Huawei Technologies Ltd., China
\and State Key Laboratory of Communication Content Cognition, People's Daily Online, China\\[0.5em]
\email{\{yaozhiyuan,zhixiao.zheng\}@mail.ustc.edu.cn, \{fzr,zdmao\}@ustc.edu.cn}\\
\email{\{jiajun.work,xssg.tuyi\}@huawei.com}
}

\maketitle

\begin{abstract}
Multimodal Large Language Models (MLLMs) are critically hampered by hallucination---generating content inconsistent with the provided image. 
In this paper, we identify an internal signature of hallucination: progressive degradation of text-to-image cross-attention during generation, leading to specific failure patterns like unfocused or biased attention. Existing mitigation strategies are largely outcome-driven and do not explicitly target this failure mode.
To address this problem, we propose {ADAPT (Attention Dynamics Alignment with Preference Tuning)}, an attention-based framework that intervenes directly on text-to-image cross-attention dynamics.
We propose ADAPT with three key contributions: a cross-attention visual anchor refined from early decoding to provide stable spatial grounding, an attention-supervised inference mechanism that detects and corrects attention drift online, and a Visual Attention Guidance DPO that aligns preferences toward visually grounded responses.
Experiments show that each component of ADAPT contributes to hallucination reduction, and the full framework achieves new best results across multiple hallucination benchmarks, reducing hallucination rates by 40–60\% across mainstream backbones while preserving general multimodal capabilities.
Our work provides an attention-based perspective on mitigating hallucinations by exploring the model's internal text-to-image cross-attention behaviors. Code is available at: \url{https://github.com/yao-ustc/ADAPT}.
\end{abstract}

\section{Introduction}
\label{sec:intro}

Multimodal Large Language Models (MLLMs) often produce {hallucinations}, which means generating descriptions or answers that are not grounded in the input image \cite{liu2024survey, bai2024hallucination}. 
Hallucinations can take the form of asserting the presence of objects that do not exist, attributing wrong properties such as color or count, or inventing relationships that are not visually supported, which severely undermines the trustworthiness of MLLMs.
This limitation persists even in commonly used models such as GPT-5 \cite{singh2025openaigpt5card}, Qwen-VL \cite{bai2025qwen2}, and LLaVA \cite{liu2023llava}. 

\begin{figure}[!t]
\centering
\includegraphics[width=\columnwidth]{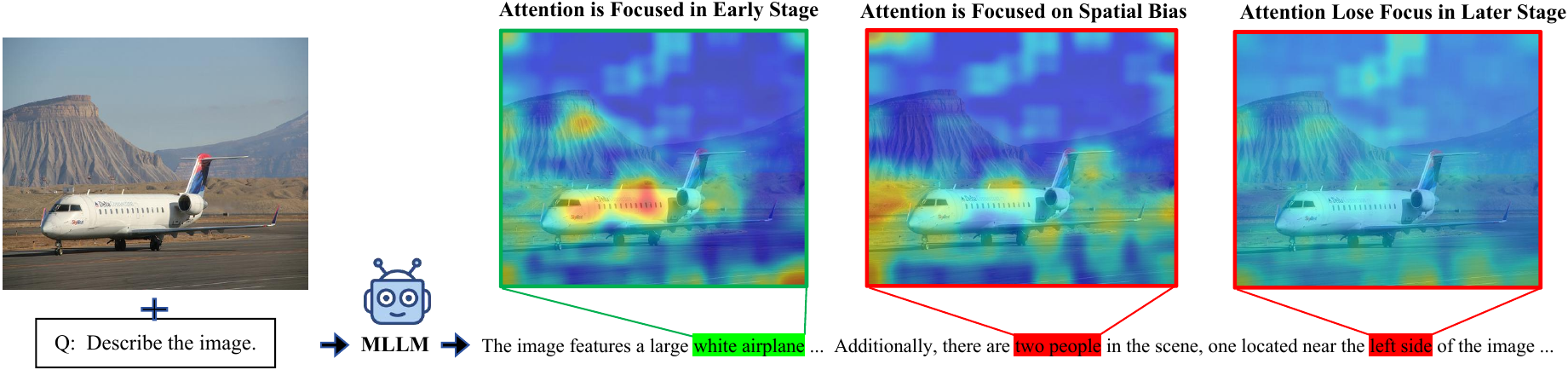}
\caption{
\textbf{Cross-attention degradation is associated with hallucinations.}
Text-to-image cross-attention is first aligned with query-relevant evidence (the airplane), then collapsing onto spatially biased regions (the left side of the airplane) and later becoming diffuse (across the scene), where hallucinated tokens appear.
}
\vspace{-6pt}
\label{fig:mechanism_analysis}
\end{figure}

\begin{figure}[!t]
\small
\centering
\includegraphics[width=\columnwidth]{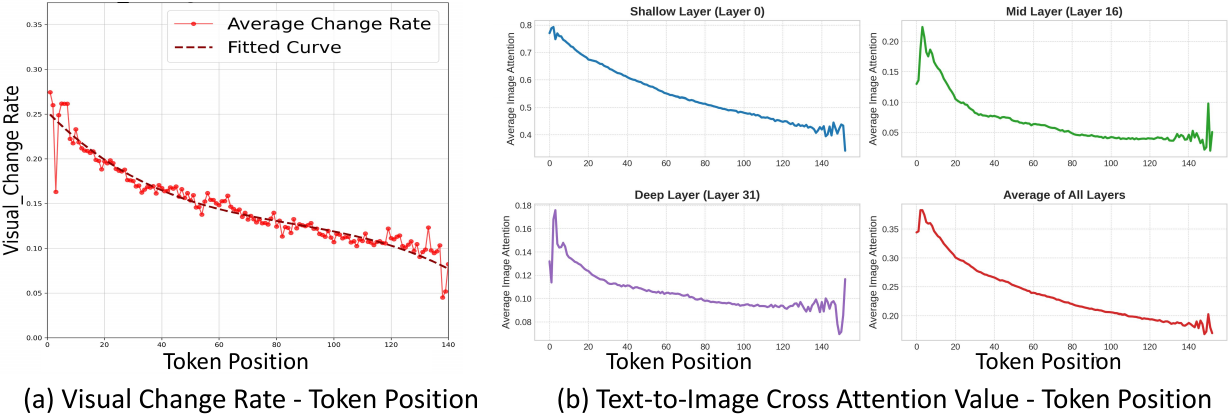}
\caption{
\textbf{Cross-attention degradation during generation.}
(a) The cross-attention change rate to newly generated text decreases as generation proceeds, dropping from about 25\% to below 10\%. 
(b) The average attention value on visual tokens decays across layers as decoding proceeds, with more than a 40\% reduction across layers, reflecting reduced reliance on visual evidence. 
}
\vspace{-12pt}
\label{fig:attention_decay}
\end{figure}

Current efforts to mitigate MLLM hallucination mainly concentrate on two directions: inference-time interventions and training-time alignment.
Inference-time methods apply plug-and-play control signals during generation, including contrastive decoding \cite{leng2024mitigating}, ensemble decoding \cite{cho2025you}, and penalty-based strategies \cite{huang2024opera}. 
These methods primarily adjust the model outputs by manipulating token-level probabilities during decoding, aiming to steer generation toward more faithful responses.
Training-time alignment methods improve visual faithfulness by updating model parameters through fine-tuning objectives, ranging from RLHF \cite{christiano2017deep} to more efficient alternatives such as DPO \cite{rafailov2023direct}, with various vision-aware variants \cite{yu2024rlaifv,wang2024mdpo,zhao2023beyond,xie2024v}. 
These methods strengthen preference alignment by shaping the model's output distribution toward human preferences for faithful descriptions.
Overall, inference-time methods adjust token-level logits and training-time alignment supervises token predictions, both of which are outcome-driven.
In contrast, internal state dynamics such as text-to-image cross-attention are closely tied to hallucinations, yet this connection is less studied.

Therefore, we analyze the model's text-to-image cross-attention as a key internal signal for tracing hallucinations.
We find that cross-attention first anchors on salient, text-relevant visual evidence and then gradually degrades during generation, shifting the model toward language priors and raising the probability of hallucinated outputs.
This degradation manifests in three aspects.
(1) Cross-attention drifts away from text-relevant visual evidence and becomes unfocused or spatially biased (Fig.~\ref{fig:mechanism_analysis}).
(2) Cross-attention changes less in response to newly generated text, with a decreasing change rate as generation proceeds (Fig.~\ref{fig:attention_decay}a).
(3) Cross-attention value on visual tokens drops across layers, indicating reduced reliance on visual evidence (Fig.~\ref{fig:attention_decay}b).
Together, these three signals show that cross-attention becomes less aligned with visual evidence, less responsive to new tokens, and less weighted on image tokens as generation proceeds, making hallucinations increasingly probable in the later part of the output (Fig.~\ref{fig:attention_decay2}).

To address this issue, we first compute a reliable, semantics-aware cross-attention anchor from early grounded stage and then use it as a supervision signal to monitor and steer cross-attention in later prior-dominated stage, maintaining semantic consistency between the generation and the image evidence. 
Building on this design, we propose \textbf{ADAPT (Attention Dynamics Alignment with Preference Tuning)}, an attention-anchor-based framework that mitigates attention degradation during long-form generation.
Our framework consists of three synergistic stages. We first perform {Visual Enhance} by extracting early, reliable text-to-image cross-attention, refining it into an anchor $A_{\text{anchor}}$. We then apply {Attention-Supervised Inference} by monitoring cross-attention against the anchor and using sparse corrective steering when attention degrades and becomes unfocused or spatially biased. Finally, we perform {Visual Attention Guidance DPO} by tuning preferences under different visual evidence conditions, using anchor-enhanced images as chosen inputs and noise images as rejected inputs to encourage visually grounded outputs.
Therefore, ADAPT refines multi-layer text-to-image cross-attention into anchors and couples anchor-guided attention supervision with preference tuning.

Our contributions are fourfold:
(1) We systematically reveal the two-stage attention degradation process behind MLLM hallucinations, identifying key failure patterns.
(2) We propose a cross-attention refinement method that uses three complementary criteria with positional bias correction to produce reliable cross-attention anchors for downstream supervision and alignment.
(3) We design an attention supervision based framework {ADAPT}, which performs online attention drift correction during inference and reinforces grounding through visual cross attention guidance preference alignment.
(4) Experiments show that ADAPT consistently outperforms strong baselines across benchmarks and backbones.

\section{Related Work}
\label{sec:related_work}

\subsection{Hallucination of MLLMs}

\begin{wrapfigure}{r}{0.50\columnwidth}
\vspace{-6pt} 
\centering
\includegraphics[width=\linewidth]{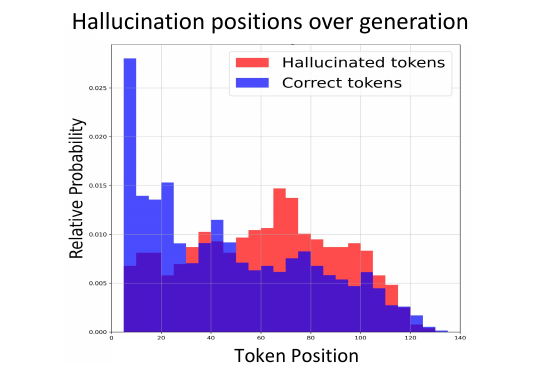}
\caption{\textbf{Hallucination positions over generation.} We compute the relative probability of hallucinatory and correct tokens across token positions and find that hallucination probability increases markedly in the later stage of generation.}
\label{fig:attention_decay2}
\vspace{-6pt} 
\end{wrapfigure}

Hallucination occurs when MLLM outputs are inconsistent with the input image \cite{liu2024survey, bai2024hallucination}. 
Existing mitigation methods mainly follow two directions: {training-time alignment} and {inference-time interventions}.

\textbf{Training-time alignment.}
Preference optimization adapts RLHF /DPO-style objectives to align model outputs with visual facts, including DPO \cite{rafailov2023direct} and vision-aware variants such as mDPO \cite{wang2024mdpo}, CLIP-DPO \cite{ouali2024clip}, SIMA \cite{wang2025enhancing}, and RLHF-V \cite{yu2024rlhf}. 
HDPO \cite{fuyuhan2025mitigatinghall} further constructs hallucination-targeted preference pairs to better cover diverse failure patterns.

\textbf{Inference-time interventions.}
Inference-time methods reduce hallucinations in a plug-and-play manner by modifying decoding \cite{lyu2024alleviating, li2025mitigating}. 
Contrastive decoding improves specificity by comparing alternative distributions \cite{leng2024mitigating}. 
IBD \cite{zhu2025ibd} amplifies image-biased attention to counter language-prior over-reliance, while SID \cite{huo2024self} introduces token-level disturbances for efficient suppression. 
DeCo \cite{wang2025mllm_can_see} leverages intermediate signals for dynamic correction during generation, and Layer Contrastive Decoding \cite{tong2025layercd} contrasts layer-wise predictions to suppress hallucinated outputs.

\subsection{Cross-attention in MLLMs}

Diagnostic studies connect anomalous attention behaviors, such as attention sinks, with failures of visual grounding \cite{jiang2025devils, kang2025see}. 
Accordingly, attention-based interventions manipulate attention heads or calibrate attention to encourage visual focus. 
For example, SPIN \cite{sarkar2025mitigating} suppresses heads with low cumulative attention to image tokens, and AVISC \cite{woo2025don} explores attentional calibration for improved faithfulness. 
Vision-guided attention methods further steer the model toward informative regions during generation \cite{zhao2025tellmodellookmitigating}.

ViT-based MLLMs also exhibit attention decay to visual tokens and positional biases over long sequences. 
CCA \cite{xing2024mitigating} reduces effective visual--text distance via token rearrangement and causal masking, FarSight \cite{tang2025seeing} improves masking and positional handling for long-range reasoning, and MemVR \cite{zou2024look} re-injects visual tokens as key--value memory under uncertainty. 
Beyond architecture-level changes, data and signal augmentation can further strengthen grounding, such as phrase-level alignment with augmented data \cite{sarkar2025dpa}.


\section{Cross-Attention and Hallucinations}
\label{sec:analysis}
We first conducted an in-depth exploration of the text-to-image cross-attention behaviors by which MLLMs generate hallucinations. Through meticulous observation of the model's attention behavior and output content, we identified two distinct stages in the generation process and three recurring degradation signals that occur when hallucinations arise.

\subsection{Cross-Attention Degradation}
We observe a two-stage shift in visual dependency during generation, from an early {Visually Grounded} stage to a later {Prior-Dominated} stage, reflected by progressive degradation of text-to-image cross-attention. This degradation manifests in three aspects:

\noindent(1) Cross-attention {progressively} drifts away from text-relevant visual evidence and becomes unfocused or spatially biased, meaning that attention is either spread broadly across many visual tokens or concentrated on irrelevant positions. In both cases, next-token prediction is {empirically} no longer constrained by query-relevant regions, so visually unsupported tokens can be selected even when the image provides contrary evidence (Fig.~\ref{fig:mechanism_analysis}).

\noindent(2) Cross-attention changes less in response to newly generated text. This indicates that attention updates are no longer driven by new context, and the model fails to re-locate when later tokens require different image regions (Fig.~\ref{fig:attention_decay}a).

\noindent(3) The overall attention value on visual tokens drops rapidly across layers and token positions. This redistribution weakens the visual constraint on decoding and makes language priors more influential (Fig.~\ref{fig:attention_decay}b).

Together, these signals indicate cross-attention degradation, mark the shift to the prior-dominated stage, and explain why hallucinations become more frequent in the later portion of the output (Fig.~\ref{fig:attention_decay2}).

\begin{figure*}[!t]
\centering
\includegraphics[width=\textwidth]{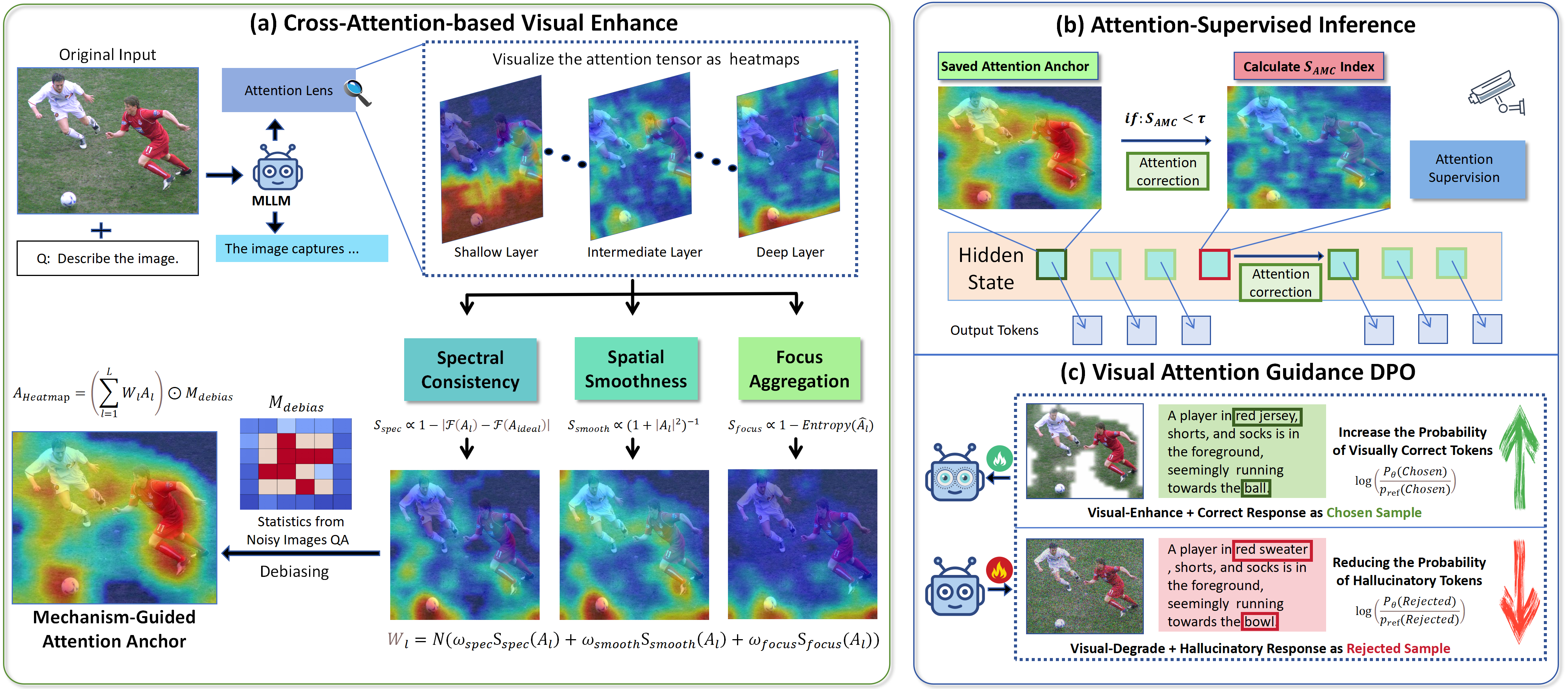} 
\caption{
\textbf{Overview of ADAPT.}
\textbf{(a) Cross-Attention-based Visual Enhance.} We extract cross-attention from the first $K$ tokens, refine it with three complementary modules, and apply debiasing to obtain a stable attention anchor.
\textbf{(b) Attention-Supervised Inference.} We compute $S_{\text{AMC}}$ at each step; when $S_{\text{AMC}}<\tau$, we trigger sparse attention correction.
\textbf{(c) Visual Attention Guidance DPO (VAG DPO).} We construct preference pairs under different visual signal strengths, using anchor-enhanced inputs as chosen samples and noise inputs as rejected samples, encouraging visually grounded generation.
}
\label{fig:framework}
\end{figure*}

\subsection{Limitations of Native Cross-Attention}
\label{sec:intrinsic_issues}

Beyond the two-stage degradation and its failure patterns, we identify a structural limitation of native text-to-image cross-attention in Transformer-based MLLMs: {attention sinks} \cite{jiang2025devils,kang2025see}. 
In AMBER, we observe that even when the visual input is uninformative (e.g., pure black/white images or heavily corrupted noise), cross-attention can still become sharply concentrated. 
This concentration is therefore not necessarily evidence-driven; instead, it often collapses onto a few recurring tokens/positions, revealing a content-independent positional bias. 
Such sink-induced bias can falsely signal high visual confidence and consequently misguide autoregressive decoding, making native cross-attention unreliable as a grounding signal.

Motivated by this limitation, we explicitly \emph{debias} cross-attention before using it for supervision or steering. 
Specifically, we estimate a model-intrinsic cross-attention bias map from noise images (Eq.~\ref{eq:Mbias}), derive a debiasing mask to suppress sink-prone positions (Eq.~\ref{eq:Mdebias}), and apply it to obtain a refined anchor (Eq.~\ref{eq:anchor_final}). 
This correction reduces spurious concentration on sink tokens, yielding anchors that better reflect query-relevant visual evidence and enabling more reliable supervision and generation.

\section{Methodology}
\label{sec:adapt_framework}

\begin{figure}[t]
    \centering
    \small
    \includegraphics[width=\columnwidth]{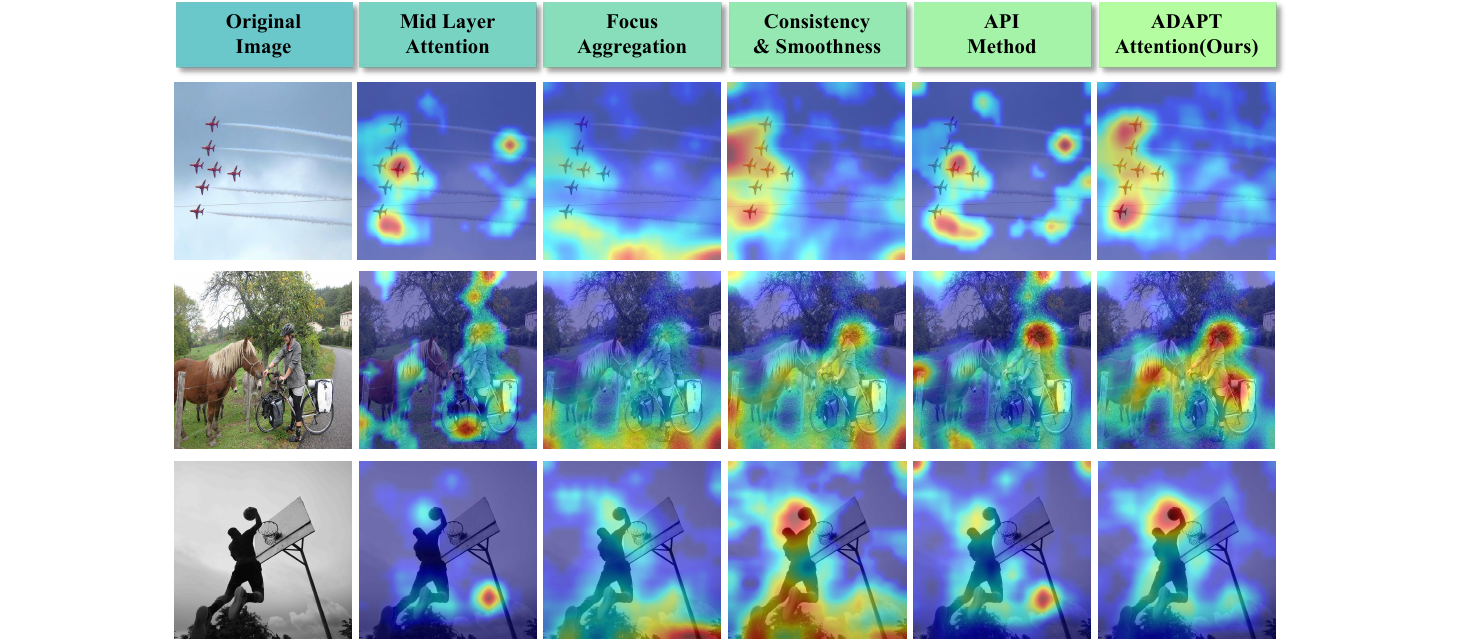}
    \caption{
\textbf{Qualitative comparison of cross-attention anchors} generated by ADAPT against various baselines, including strategies in Section~\ref{ASI} and the API method \cite{yu2024attention}. All heatmaps are obtained by fusing multi-layer query-to-image cross-attention.
    }
    \label{fig:case_study_2}
\end{figure}

Building on the Cross-Attention Visual Anchor, we propose \textbf{ADAPT (Attention Dynamics Alignment with Preference Tuning)}, an attention driven framework that uses a shared visual anchor across coordinated components, as shown in Fig.~\ref{fig:framework}. ADAPT intervenes at three stages: a lightweight pre-inference visual enhancement step, an inference-time attention supervision module, and a training-time preference optimization module. Among them, the core interventions are {Attention-Supervised Inference} and {Visual Attention Guidance DPO}, which address hallucination during decoding and alignment.


\subsection{Cross-Attention-based Visual Enhance}
\label{VE}

Motivated by Section~\ref{sec:analysis}, we treat text-to-image cross-attention as a controllable internal signal for hallucination mitigation. Since raw cross-attention is noisy and position-biased, and a reliable reference must be query-dependent, we refine early multi-layer cross-attention into a {Cross-Attention Visual Anchor} via complementary criteria and explicit debiasing, yielding a stable, query-relevant anchor for supervision and preference tuning.

To obtain a stable, factually grounded reference, we construct a {cross-attention visual anchor} from early-generation cross-attention. Let $\mathcal{A}=\{A_1,\dots,A_L\}$ be the 2D cross-attention maps from $L$ Transformer layers, where $A_l\in\mathbb{R}^{H\times W}$ is conditioned on the query $Q$ and the initial prefix. We score each $A_l$ with three criteria targeting common failures—noise, unfocused attention, and query-mismatched concentration ${S}(A_l,Q)\to[0,1]^3$, yielding $\mathbf{s}_l=[S_{\text{spec}},S_{\text{smooth}},S_{\text{focus}}]^\top$, and fuse the maps accordingly.

\subsubsection{Dynamic Fusion Weight Calculation.}
\label{sec:dynamic_fusion}
Unreliable attention maps often exhibit excessive high-frequency components, which correspond to visually noisy and unstable grounding.
We quantify this property using the high-frequency energy ratio:
{\small
\begin{equation}
\Psi(A_l)=\frac{\sum_{(u,v)\in\Omega_H}\left|\mathcal{F}(A_l)_{u,v}\right|^2}{\sum_{u,v}\left|\mathcal{F}(A_l)_{u,v}\right|^2+\epsilon},
\end{equation}
where $\mathcal{F}(\cdot)$ is the 2D FFT and $\Omega_H$ denotes the high-frequency band.
We then score deviation from a layer-wise reference $\mu_l^{\text{spec}}$ :
\begin{equation}
S_{\text{spec}}(A_l)=\exp\!\left(-\gamma_1\left|\Psi(A_l)-\mu_l^{\text{spec}}\right|\right).
\end{equation}}

To suppress fragmented or scattered activations (the {unfocused} pattern) and favor contiguous object regions, we compute an inverse TV regularizer on the normalized map $\hat{A}_l$:
{\small
\begin{equation}
S_{\text{smooth}}(A_l)=\left(1+\lambda_{\text{TV}}\frac{1}{HW}\sum_{i,j}\|\nabla \hat{A}_l(i,j)\|_2\right)^{-1}.
\end{equation}}

Over-diffuse attention may ignore key evidence, while over-concentrated attention can miss required context; importantly, the desired focus depends on the query.
Let $\bar{H}(\hat{A}_l)$ denote the normalized Shannon entropy of $\hat{A}_l$. 
We define a query-dependent target entropy
\begin{equation}
\bar{H}^*(Q)=(1-g(Q))\bar{H}_{\text{obj}}+g(Q)\bar{H}_{\text{global}},
\end{equation}
and score by deviation:
{\small
\begin{equation}
S_{\text{focus}}(A_l,Q)=1-\left|\bar{H}(\hat{A}_l)-\bar{H}^*(Q)\right|.
\end{equation}}

After obtaining $\mathbf{s}_l$, we compute layer-wise fusion weights via a linear projection followed by Softmax normalization:
{\small
\begin{equation}
W_l=\frac{\exp(\boldsymbol{\omega}^\top \mathbf{s}_l)}{\sum_{k=1}^{L}\exp(\boldsymbol{\omega}^\top \mathbf{s}_k)},
\end{equation}}
where $\boldsymbol{\omega}=[\omega_{\text{spec}},\omega_{\text{smooth}},\omega_{\text{focus}}]^\top$ controls the relative importance of the criteria.
This emphasizes layers whose attention better matches the above grounding properties.
The fused attention map is $\sum_{l=1}^{L}W_l\cdot A_l$.

\subsubsection{Spatial Bias Correction.}
\label{sec:spatial_bias_correction}

Transformer-based MLLMs can exhibit content-independent positional priors, where certain spatial tokens are systematically over-attended regardless of image content (the {biased} pattern). 
To estimate this {content-free} bias, we feed noise images while keeping the same questions and average the resulting fused attention maps:
\footnotesize
\begin{equation}
M_{\text{bias}}=
\mathbb{E}_{(I,Q)\sim \mathcal{D}_{\text{SVA}}}
\left[
\mathrm{Normalize}\!\left(A_{\text{fused}}(I_{\text{noise}},Q)\right)
\right],
\label{eq:Mbias}
\end{equation}
\normalsize
where $\mathrm{Normalize}(\cdot)$ rescales the map to sum to $1$.
We then construct a debiasing mask by applying an element-wise inverse (with smoothing) followed by renormalization:
\small
\begin{equation}
M_{\text{debias}}=
\mathrm{Normalize}\!\left((M_{\text{bias}}+\epsilon)^{-1}\right).
\label{eq:Mdebias}
\end{equation}
\normalsize
Finally, we apply the debiasing mask $M_{\text{debias}}$ to the fused multi-layer cross-attention map via element-wise multiplication, removing layer-aggregated positional bias and yielding the debiased visual anchor:
{\small
\begin{equation}
A_{\text{anchor}}=
\left(\sum_{l=1}^{L} W_l \cdot A_l\right)\odot M_{\text{debias}}.
\label{eq:anchor_final}
\end{equation}}

The resulting anchor  is conditioned on the query semantics and shifts to different image regions as the question changes. It is also layer-selective and corrected for positional bias, and is used for both inference-time attention supervision and preference-pair construction.

\subsubsection{Hyperparameter Selection for Fusion Weights.}
\label{sec:weight_selection}
We select $\boldsymbol{\omega}$ via a constrained grid search on AMBER with $\omega_k\ge 0$ and $\sum_k \omega_k=1$ using the Chair metric, and use the same default across backbones:
$\omega_{\text{spec}}=0.4,\ \omega_{\text{smooth}}=0.3,\ \omega_{\text{focus}}=0.3$.
As visualized in Fig.~\ref{fig:two_in_one_row}(a), the 3D surface over the weight simplex exhibits a broad low-score region, which allows the hyperparameter choice to be inspected directly and indicates low sensitivity around the selected setting.
Moreover, the optimum does not collapse to a single criterion weight, suggesting that the three refinement criteria provide complementary benefits in forming a reliable cross-attention anchor.
Full results are provided in the Appendix.

\subsubsection{Anchor-Guided Visual Enhancement.}
As a lightweight pre-inference step, we use the visual anchor to create a visually grounded input that reduces the risk of later attention drift. Specifically, we first perform a short preliminary inference run (the first $K$ tokens) during the reliable Strong Visual Anchoring Phase to obtain the anchor $A_{\text{anchor}}$. We apply the attention heatmap and a mild binary mask to the original image, producing an anchor-guided image that remains visually interpretable for manual inspection of anchor quality. During extended QA processes, critical visual anchors are generated first; these anchors are subsequently fed back into the model as Visual Enhance to guide the rest of the response generation. Furthermore, these masked images are stored for later training, providing the model with explicit visual cues about which regions should be treated as important.

\subsection{Attention-Supervised Inference}
\label{ASI}

To maintain visual grounding as decoding transitions to the {Prior-Dominated} stage, we introduce {Attention-Supervised Inference}. While hallucinations may also stem from other factors, we treat attention degradation as a practical, model-internal signal for grounding. Rather than forcing cross-attention to strictly follow a fixed anchor, our module monitors attention fidelity online and intervenes only when substantial degradation is detected, applying sparse corrective steering to recover from unfocused or spatially biased drift.

At each decoding step $t$, we evaluate the alignment between the current cross-attention distribution $A_t \in \mathbb{R}^N$ and the previously synthesized visual anchor $A_{\text{anchor}}$. Importantly, we use the anchor as a {reference} for {detection}, not as a hard target: the model is free to adapt attention as the textual context evolves, and we intervene only when attention drifts toward unfocused or spatially biased attention. We define the {Anchor-Modulated Concentration (AMC)} index ${S}_{\text{AMC}}$ as an online attention-fidelity score for triggering such selective corrections:
{\small
\begin{equation}
    {S}_{\text{AMC}}(A_t \,\|\, A_{\text{anchor}}) = \frac{\sum_{i=1}^{N} (a_{i,t}^2 \cdot w_{i}^{\text{anchor}})}{\left( \sum_{i=1}^{N} a_{i,t} \right)^2 + \epsilon}
\end{equation}}

where $a_{i,t}$ is the attention weight of the $i$-th image token at step $t$, and $w_{i}^{\text{anchor}}$ is its anchor weight. AMC combines concentration ($a_{i,t}^2$) with anchor consistency (down-weighting regions with near-zero $w_{i}^{\text{anchor}}$), enabling online drift detection during autoregressive decoding.

We monitor attention fidelity during decoding. When ${S}_{\text{AMC}}<\tau$, we apply a steering operator $\mathcal{O}$ to the attention logits:
{\small
\begin{equation}
    \hat{a}_{i,t} =
    \begin{cases}
      a_{i,t} + \alpha \log(w_i^{\text{anchor}} + \epsilon), & \text{if } {S}_{\text{AMC}} < \tau \\
      a_{i,t}, & \text{otherwise}
    \end{cases}
\end{equation}}
We set $\tau=0.6$ using a validation sweep to select the Pareto-optimal trade-off between hallucination reduction and runtime overhead (Appendix).

\begin{figure*}[!t]
\tiny
\centering
\includegraphics[width= \textwidth]{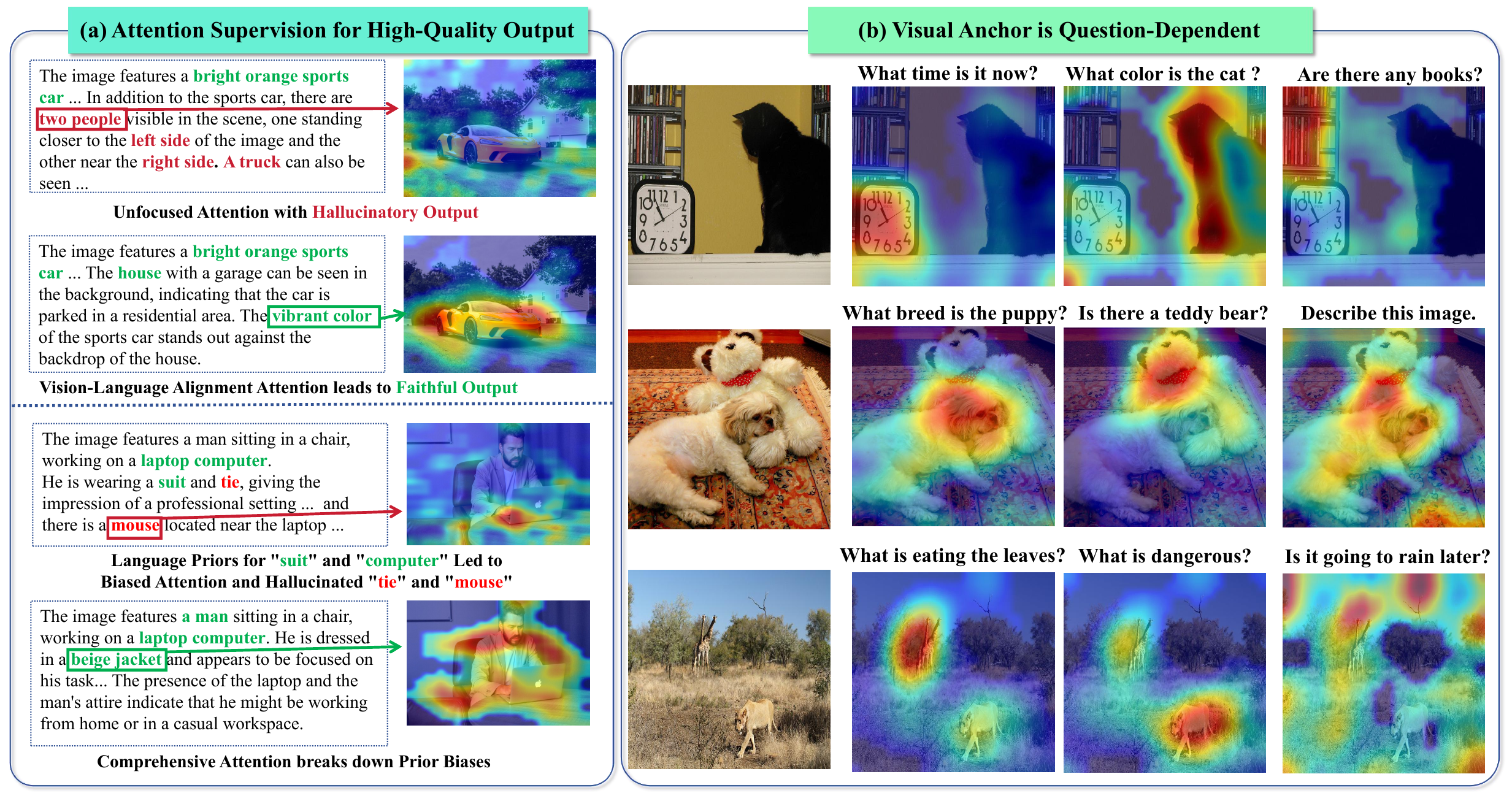}
\caption{
\textbf{(a) Qualitative examples of attention-supervised inference}. We visualize how the refined cross-attention anchor enables attention correction and removes hallucination.
\textbf{(b) Query-dependent cross-attention anchors.} The refined cross-attention anchor shifts to different image regions depending on the question semantics.
}
\label{fig:qualitative_examples}
\end{figure*}

\subsection{Visual Attention Guidance DPO}
\label{VAG-DPO}

While Attention-Supervised Inference mitigates hallucination at inference time, we further improve the model's intrinsic preference for visually grounded responses through {VAG-DPO}. Unlike standard DPO, which contrasts responses primarily at the textual level, VAG-DPO contrasts responses under different {visual evidence conditions}, thereby explicitly decoupling visual grounding from linguistic priors.

We construct preference pairs as follows:
\begin{itemize}
    \item \textbf{Chosen sample $(I_{\text{VE}}, y_c)$:} The factual response $y_c$ is conditioned on the {visually enhanced} image $I_{\text{VE}}$, which is derived from the Cross-Attention Visual Anchor. This strengthens the association between correct tokens and salient visual evidence.
    \item \textbf{Rejected sample $(I_{\text{noise}}, y_r)$:} The hallucinatory response $y_r$ is paired with a {noise-injected} image $I_{\text{noise}}$, creating a ``blind'' condition in which the model should not assign high confidence to visually unsupported content.
\end{itemize}

Formally, we optimize the following objective:
\footnotesize
\begin{equation}
    \mathcal{L}_{\text{VAG-DPO}} = -\mathbb{E}_{\mathcal{D}} \left[ \log \sigma \left( \beta \log \frac{P_\theta(y_c \mid I_{\text{VE}}, Q)}{P_{\text{ref}}(y_c \mid I_{\text{VE}}, Q)} - \beta \log \frac{P_\theta(y_r \mid I_{\text{noise}}, Q)}{P_{\text{ref}}(y_r \mid I_{\text{noise}}, Q)} \right) \right]
\end{equation}
\normalsize

By explicitly distinguishing the visual evidence conditions ($I_{\text{VE}}$ versus $I_{\text{noise}}$), VAG-DPO penalizes the model's tendency to generate high-confidence hallucinations from language priors alone, and encourages generation probabilities to remain contingent on valid visual support.

\section{Experiments}

\begin{table*}[t]
\caption{\textbf{Main results.} Performance comparison on multiple hallucination benchmarks. {ADAPT-TF} denotes our training-free variant with Visual Enhance and Attention-Supervised Inference, while full {ADAPT} adds VAG-DPO. Full ADAPT achieves the best overall performance among compared methods across backbones.}
\label{tab:main_results}

\scriptsize
\centering
\renewcommand{\arraystretch}{1.1}
\setlength{\tabcolsep}{1mm}
\begin{tabular}{@{}l cc cccc cc c@{}} 
\toprule
\multirow{2}{*}{\textbf{Methods}} & 
\multicolumn{2}{c}{Obj-Hal \cite{rohrbach2018object}} & 
\multicolumn{4}{c}{AMBER \cite{wang2023amber}} & 
\multicolumn{2}{c}{MM-Hal \cite{sun2023aligning}} & 
\multicolumn{1}{c}{POPE-Adv \cite{li2023evaluating}} \\
\cmidrule(lr){2-3} \cmidrule(lr){4-7} \cmidrule(lr){8-9} \cmidrule(lr){10-10} 
& Chair$_s$$\downarrow$ & Chair$_i$$\downarrow$ 
& Chair$\downarrow$ & Hal$\downarrow$ & Cog$\downarrow$ 
& Acc.$\uparrow$ 
& Score$\uparrow$ & Hal$\downarrow$ 
& Precision$\uparrow$ \\
\midrule

LLaVA-v1.5-7B \cite{liu2023improved}      & 54.4 & 15.8 & 7.8 & 36.4 & 4.2 & 71.7 & 1.86 & 0.64 & 76.1 \\
\ \ +DPO \cite{rafailov2023direct}      & 49.0 & 13.0 & 6.5 & 34.5 & 2.3 & 71.3 & 2.14 & 0.65 & 77.8 \\
\ \ +mDPO \cite{wang2024mdpo}           & 35.7 & 9.8 & 4.4 & 24.5 & 2.4 & 71.1 & 2.39 & 0.54 & 77.5 \\
\ \ +SIMA \cite{wang2025enhancing} & 40.9 & 10.4 & - & - & - & - & 2.30 & - & - \\
\ \ +VCD \cite{leng2024mitigating}      & 51.2 & 15.0 & 7.4 & 34.7 & 4.1 & 71.8 & 2.12 & 0.54 & 73.4 \\
\ \ +OPERA \cite{huang2024opera}       & 50.8 & 14.3 & 6.9 & 30.3 & 3.2 & 75.2 & 2.15 & 0.54 & 82.2 \\
\ \ +AVISC \cite{woo2025don} & - & - & 6.25 & 25.6 & 2.0 & - & - & - & 79.8 \\
\rowcolor[gray]{0.92}
\ \ +\textbf{ADAPT-TF} & 35.6 & 11.8 & 4.0 & 16.7 & \textbf{1.1} & 73.5 & 2.55 & 0.53 & 90.3 \\
\rowcolor[gray]{0.92}
\ \ +\textbf{ADAPT} & \textbf{30.7} & \textbf{8.5} & \textbf{3.8} & \textbf{15.2} & 1.2 & \textbf{79.6} & \textbf{2.62} & \textbf{0.53} & \textbf{91.9} \\

\midrule
LLaVA-v1.5-13B \cite{liu2023improved} & 49.8 & 14.6 & 7.0 & 29.0 & 3.2 & 71.4 & 2.38 & 0.53 & 84.5 \\
\ \ +DPO \cite{rafailov2023direct}      & 43.9 & 12.4 & 6.1 & 26.3 & 2.7 & - & 2.47 & 0.51 & 82.7 \\
\ \ +RLHF-V \cite{yu2024rlhf}      & - & - & - & - & - & 72.6 & 2.50 & 0.52 & - \\
\ \ +VCD \cite{leng2024mitigating}      & 53.6 & 15.3 & 7.2 & 28.7 & 3.3 & 73.5 & 2.31 & 0.54 & 82.4 \\
\ \ +OPERA \cite{huang2024opera}       & 42.6 & 13.2 & 6.8 & 28.5 & 3.1 & - & 2.46 & 0.52 & 84.7 \\
\rowcolor[gray]{0.92}
\ \ +\textbf{ADAPT-TF} & 35.4 & 11.6 & \textbf{4.0} & 18.0 & 1.5 & \textbf{82.4} & 2.55 & 0.51 & 90.6 \\
\rowcolor[gray]{0.92}
\ \ +\textbf{ADAPT } & \textbf{34.8} & \textbf{10.6}& 4.2 & \textbf{17.7} & \textbf{1.4} & \textbf{82.4} & \textbf{2.64} & \textbf{0.49} & \textbf{90.6} \\

\midrule
Qwen2.5-VL-3B \cite{bai2025qwen2} & 43.2 & 12.2 & 8.0 & 44.4 & 4.3 & 86.0 & 2.62 & 0.49 & 93.2 \\
\ \ +VCD \cite{leng2024mitigating} & 41.4 & 12.3 & 7.7 & 40.8 & 4.2 & 85.3 & 3.14 & 0.42 & 92.9 \\
\ \ +DPO \cite{rafailov2023direct} & 30.6 & 10.7 & 5.7 & 22.6 & 1.4 & 82.8 & 2.53 & 0.52 & 93.0 \\
\rowcolor[gray]{0.92}
\ \ +\textbf{ADAPT-TF } & 19.8 & 8.1 & 6.2 & 23.7 & 1.1 & \textbf{88.9} & 3.11 & 0.39 & 93.5 \\
\rowcolor[gray]{0.92}
\ \ +\textbf{ADAPT } & \textbf{15.2} & \textbf{5.3} & \textbf{3.1} & \textbf{13.5} & \textbf{0.5} & 88.3 & \textbf{3.16} & \textbf{0.37} & \textbf{94.0} \\

\midrule
Qwen2.5-VL-7B \cite{bai2025qwen2} & 32.3 & 9.6& 4.7 & 21.9 & 1.2 & 87.6 & 3.55 & 0.4 & 93.5 \\
\ \ +VCD \cite{leng2024mitigating} & 35.1 & 10.0 & 4.8 & 23.9 & 1.1 & 74.2 & 3.42 & 0.4 & 93.7 \\
\ \ +DPO \cite{rafailov2023direct} & 29.6 & 11.7 & 3.9 & 17.3 & \textbf{0.6} & 85.8 & 3.59 & 0.39 & 92.4 \\
\ \ +SPIN \cite{sarkar2025mitigating} & 28.6 & 7.0 & - & - & - & - & - & - & - \\
\rowcolor[gray]{0.92}
\ \ +\textbf{ADAPT-TF } & 27.8 & 8.2 & 5.3 & 19.5 & 0.8 & 86.3 & \textbf{3.83} & 0.31 & \textbf{93.9} \\
\rowcolor[gray]{0.92}
\ \ +\textbf{ADAPT } & \textbf{16.8} & \textbf{5.6} & \textbf{3.5} & \textbf{13.1} & 0.7 & \textbf{88.1} & 3.75 & \textbf{0.30} & 93.5 \\

\bottomrule
\end{tabular}
\end{table*}

\subsection{Experimental Setup}

\noindent
\textbf{Benchmark:} We evaluate our method on four widely-recognized hallucination benchmarks: {MM-Hal} \cite{sun2023aligning} for general QA; {Obj-Hal} \cite{rohrbach2018object} (using the $Chair$ metric) for object-level errors; {AMBER} \cite{wang2023amber}, providing fine-grained evaluation across {generative} and {discriminative} tasks; and {POPE} \cite{li2023evaluating} for probing object existence.

\noindent
\textbf{Compared Methods:}
We compare our approach against a suite of baselines and state-of-the-art methods. These include the basic LLaVA-v1.5 \cite{liu2023improved} and Qwen2.5-VL \cite{bai2025qwen2} model, various inference-time decoding strategies such as VCD \cite{leng2024mitigating},  OPERA \cite{huang2024opera}, ED \cite{cho2025you}, AVISC \cite{woo2025don}, SPIN \cite{sarkar2025mitigating} and GF-SCD \cite{zhang2025self}, as well as preference optimization techniques like RLHF-V \cite{yu2024rlhf}, mDPO \cite{wang2024mdpo} and SIMA \cite{wang2025enhancing}.

\noindent
\textbf{Implementation Details:}
We ensure a fair comparison for open-source methods by utilizing the exact same datasets and training parameters. Furthermore, for all GPT evaluations, we employed the consistent {gpt-4-turbo} version, with the final results reported as the average of three evaluation runs. Training parameters are detailed in the Appendix.

\noindent
\textbf{ADAPT Parameters:} We set parameters based on our analysis and empirical studies. According to the sensitivity study in Fig.~\ref{fig:two_in_one_row}a, we choose the fusion weights $\omega_{\text{spec}}=0.4$, $\omega_{\text{smooth}}=0.3$, and $\omega_{\text{focus}}=0.3$ for cross-attention refinement. Based on the ablation study in the appendix and the analysis in Section~\ref{ASI}, we generate the visual anchor from the first $K=5$ tokens and set the AMC intervention threshold to $\tau=0.6$.

\noindent
\textbf{Preference Dataset Construction:}
For the VAG-DPO Training stage, we constructed our preference dataset based on the subset of the RLAIF-V \cite{yu2024rlaifv} and RLHF-V \cite{yu2024rlhf} dataset. 
While we directly adopted the response preference pairs from the sampled dataset, the chosen image is the visual anchor image generated by our approach and the rejected image is a noise image. 
This process transforms the original text-preference dataset into a vision-conditioned preference dataset tailored to our framework.

\begin{table}[t]
\centering
\small
\caption{\textbf{Ablation and efficiency analysis.}}
\label{tab:ablation_efficiency}
\begin{subtable}[t]{0.49\linewidth}
\centering
\caption{Ablation study of ADAPT modules. Each ADAPT module improves hallucination metrics, and the full ADAPT performs best.}
\label{tab:ablation}
\setlength{\tabcolsep}{2mm}

\begin{adjustbox}{width=\linewidth}
\begin{tabular}{lccc}
\toprule
\textbf{Method} & Chair$\downarrow$ & Hal$\downarrow$ & Cog$\downarrow$ \\
\midrule
LLaVA-v1.5-7B & 7.8 & 36.4 & 4.2 \\
\midrule
+ Visual Enhance (\ref{VE}) & 4.7 & 18.7 & 1.8 \\
+ Attention-Supervised (\ref{ASI}) & 6.3 & 27.8 & 2.9 \\
+ VAG-DPO (\ref{VAG-DPO}) & 6.0 & 25.0 & 2.1 \\
\midrule
\rowcolor[gray]{0.92}
\textbf{+ ADAPT} & \textbf{3.8} & \textbf{15.2} & \textbf{1.2} \\
\bottomrule
\end{tabular}
\end{adjustbox}

\end{subtable}
\hfill
\begin{subtable}[t]{0.49\linewidth}
\centering
\caption{Inference efficiency and performance. \\
ADAPT achieves the strongest hallucination reduction with only \(1.42\times\) runtime overhead.}
\label{tab:efficiency}
\setlength{\tabcolsep}{2mm}

\begin{adjustbox}{width=\linewidth}
\begin{tabular}{@{}lccc@{}}
\toprule
\textbf{Method} & Avg. time (s)$\downarrow$ & Time Ratio$\downarrow$ & Chair$_s$$\downarrow$ \\
\midrule
LLaVA-v1.5-7B & 2.71 & 1.00 & 53.6 \\
\midrule
+ VCD \cite{leng2024mitigating} & 5.50 & 2.03 & 48.8 \\
+ OPERA \cite{huang2024opera} & 19.51 & 7.20 & 45.1 \\
+ GF-SCD \cite{zhang2025self} & 10.75 & 3.96 & 48.8 \\
\midrule
\rowcolor[gray]{0.92}
\textbf{+ ADAPT} & \textbf{3.84} & \textbf{1.42} & \textbf{26.3} \\
\bottomrule
\end{tabular}
\end{adjustbox}

\end{subtable}

\end{table}

\subsection{Main Results}

As shown in Table~\ref{tab:main_results}, our ADAPT method achieves State-of-the-Art (SOTA) performance across all mainstream hallucination benchmarks (AMBER, POPE, MM-Hal, Obj-Hal). Our core motivation was to mitigate hallucination in long-form generation, successfully achieving SOTA on long captioning tasks. Crucially, ADAPT demonstrates architectural generalizability across diverse MLLM families, validated by strong results on both LLaVA-v1.5-7B/13B and Qwen2.5-VL-3B/7B. Specifically, ADAPT establishes superior control over object errors ($Chair_s, Chair_i$) and achieves over 90\% precision on POPE Adversarial. Furthermore, our lightweight, training-free variant, ADAPT-TF, surpasses most advanced methods, proving the practical potential and overall effectiveness of our multi-stage intervention framework in mitigating model hallucinations.

\subsection{Ablation Study}

\noindent
\textbf{Hallucination Mitigation:} We evaluated the contribution of each high-level module on the AMBER dataset. As shown in Table~\ref{tab:ablation}, we started with the LLaVA-v1.5-7B baseline and progressively added our components. The results show that each module brings a notable improvement. The full framework achieves the best performance across all metrics, highlighting a strong synergistic effect and demonstrating that all three components are crucial to the result.

\begin{figure}[!t]
    \centering
    \begin{subfigure}[t]{0.49\columnwidth}
        \centering
        \includegraphics[width=\linewidth]{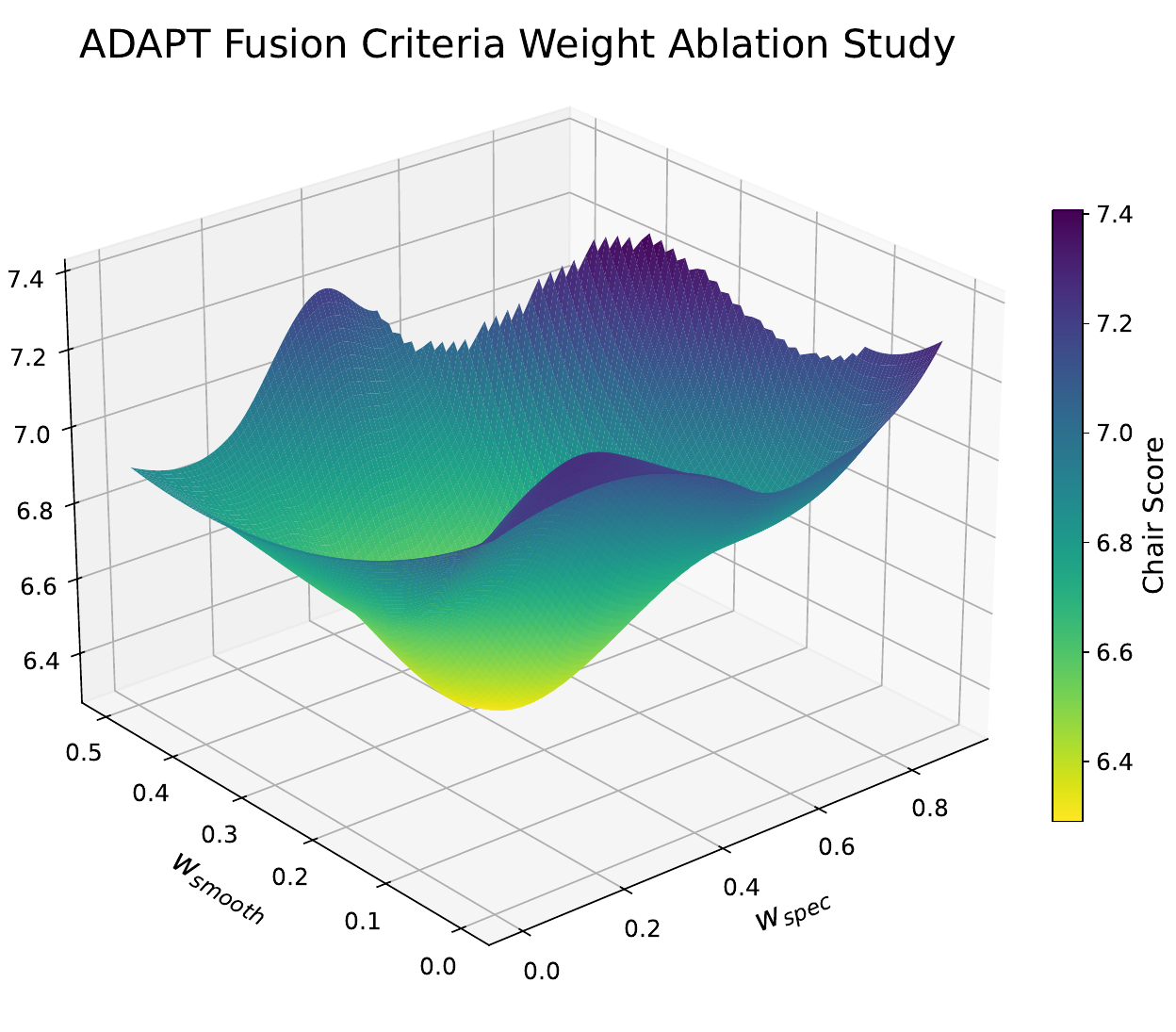}
        \caption{}
        \label{fig:fusion_weight_sensitivity}
    \end{subfigure}
    \begin{subfigure}[t]{0.49\columnwidth}
        \centering
        \includegraphics[width=\linewidth]{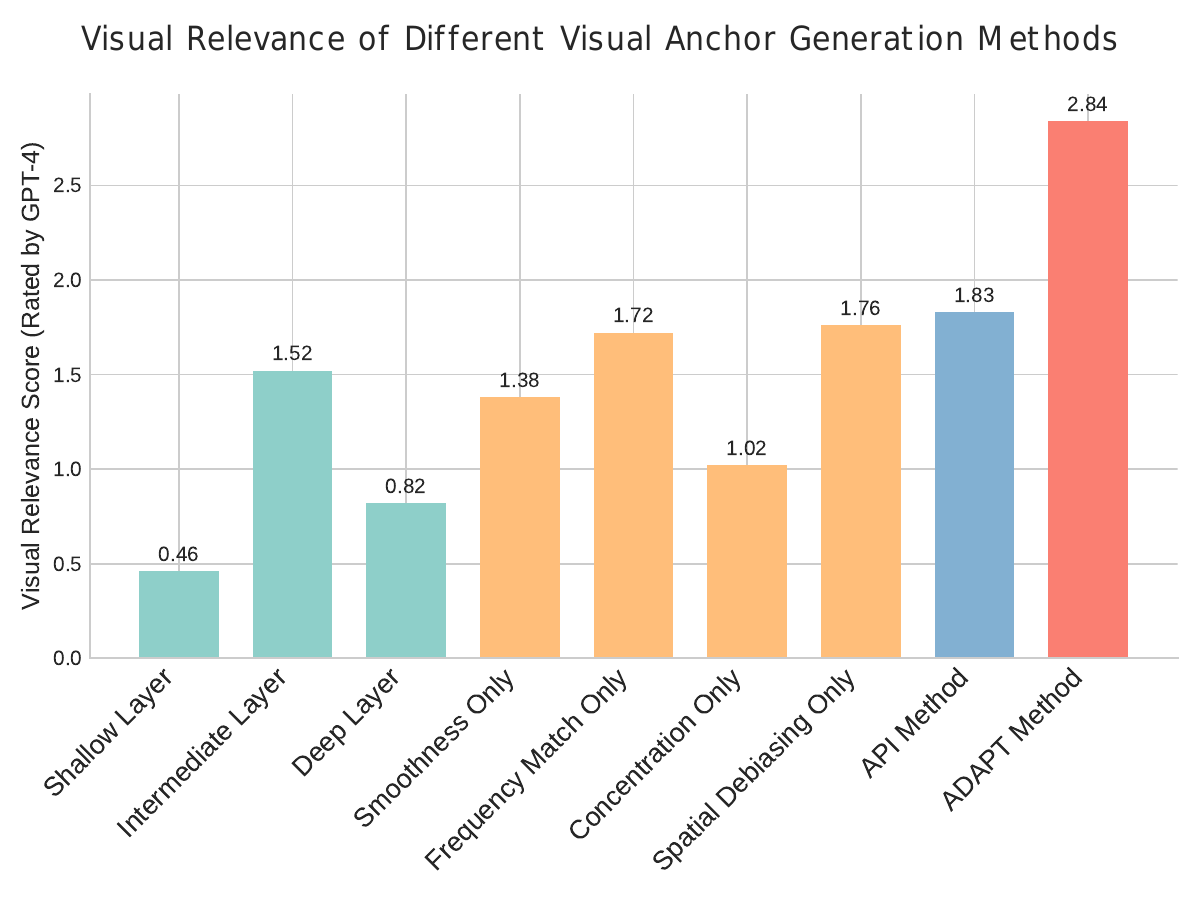}
        \caption{}
        \label{fig:heatmap_ablation}
    \end{subfigure}\hfill

    \caption{\textbf{(a)} \textbf{Fusion Weight Sensitivity of ADAPT.} 3D surface of AMBER Chair score as a function of fusion weights $w_{\text{spec}}$ and $w_{\text{smooth}}$. For Chair Score, Lower is better. \textbf{(b)} \textbf{Evaluation of Visual Anchor Semantic Relevance.} We compare our ADAPT anchor against ablated variants and API baseline; higher scores indicate better highlighting of query-relevant evidence, and ADAPT performs best.}
    \label{fig:two_in_one_row}
\end{figure}

\noindent
\textbf{Visual Anchor Quality:}
To quantitatively assess the visual relevance of our generated visual anchors, we compare our approach against several baselines: single-layer attention variants, ablated fusion criteria, and the SOTA {API method} \cite{yu2024attention}. Figure~\ref{fig:heatmap_ablation} shows our full approach significantly outperforms all baselines.

We conducted more ablation studies on both {hyperparameter selection} and {construction of preference pairs} (detailed results in Appendix).

\subsection{Further Analysis}

\noindent
\textbf{Inference Efficiency:}
We evaluate the inference efficiency of our method to demonstrate its practicality. As detailed in Table~\ref{tab:efficiency}, our ADAPT introduces a modest time overhead, requiring only 1.42 times the inference time of the regular baseline. This is significantly more efficient than others. 


\noindent
\textbf{Visual Grounding Evaluation:}
To provide a rigorous, objective validation of our {Cross-Attention Visual Anchor} quality, we assess its efficacy as a visual grounding feature.
We measure selection Accuracy (Acc) and mean Intersection over Union (mIoU) on RefCOCOg Test and RefCOCO Val datasets \cite{subramanian2022reclip} by selecting the bounding box that maximizes the contained Visual Anchor's average attention value. As shown in Table~\ref{tab:visual_anchor_grounding_results}, ADAPT achieves SOTA Acc and mIoU, outperforming baselines LLaVA-v1.5-7B \cite{liu2023improved}, Vanilla CLIP \cite{radford2021learning}, and the SOTA API method. This confirms ADAPT successfully imparts robust grounding capabilities to a non-localizing base VLM.

\noindent
\textbf{General Capability Evaluation:}
Since preference optimization can inadvertently reduce a model's general-purpose multimodal capability, we further evaluate the impact of our training stage on two widely-used general benchmarks, {MME} \cite{fu2025mmecomprehensiveevaluationbenchmark} and {MMBench} \cite{liu2024mmbenchmultimodalmodelallaround}. 
As shown in Table~\ref{tab:general_retention}, both vanilla DPO and our VAG-DPO cause a mild drop relative to the base model. This suggests that contrasting responses under different {visual evidence conditions} strengthens visual alignment while better preserving general capabilities.

\begin{table*}[t]
\centering
\scriptsize
\caption{\textbf{Further analysis of grounding and general capability.} }
\begin{subtable}[t]{0.49\textwidth}
\centering
\caption{\textbf{Evaluation of using attention for the grounding task} on RefCOCO \cite{subramanian2022reclip}. Using our visual anchor attention for localization yields {significant improvements over the LLaVA-v1.5-7B \cite{liu2023improved}} and {outperforms the Vanilla CLIP \cite{radford2021learning}}.}
\label{tab:visual_anchor_grounding_results}
\setlength{\tabcolsep}{3pt}
\renewcommand{\arraystretch}{1.0}
\begin{tabular}{lcccc}
\toprule
Method & \multicolumn{2}{c}{RefCOCOg} & \multicolumn{2}{c}{RefCOCO} \\
\cmidrule(lr){2-3}\cmidrule(lr){4-5}
& Acc$\uparrow$ & mIoU$\uparrow$ & Acc$\uparrow$ & mIoU$\uparrow$ \\
\midrule
CLIP \cite{radford2021learning} & 45.9 & 51.3 & 39.3 & 45.3 \\
7B Base \cite{liu2023improved} & 32.6 & 41.8 & 30.2 & 37.3 \\
\ \ +API \cite{yu2024attention} & 39.8 & 56.0 & 32.5 & 39.1 \\
\ \ +Focus & 27.9 & 37.0 & 25.9 & 34.9 \\
\ \ +Spectral & 42.3 & 49.8 & 37.1 & 45.6 \\
\ \ +Spatial & 37.0 & 45.1 & 32.6 & 40.5 \\
\rowcolor[gray]{0.92}
\ \ \textbf{+ADAPT} & \textbf{50.3} & \textbf{57.0} & \textbf{43.0} & \textbf{50.4} \\
\bottomrule
\end{tabular}
\end{subtable}
\hfill
\begin{subtable}[t]{0.49\textwidth}
\centering
\caption{\textbf{General capability on MME \cite{fu2025mmecomprehensiveevaluationbenchmark} and MMBench \cite{liu2024mmbenchmultimodalmodelallaround}.} 
While both DPO variants slightly reduce general performance, our VAG-DPO causes a smaller degradation than vanilla DPO, indicating better capability retention.}
\label{tab:general_retention}
\setlength{\tabcolsep}{3pt}
\renewcommand{\arraystretch}{1.0}
\begin{tabular}{lcccc}
\toprule
Method & \multicolumn{2}{c}{MME \cite{fu2025mmecomprehensiveevaluationbenchmark}} & \multicolumn{2}{c}{MMBench \cite{liu2024mmbenchmultimodalmodelallaround}} \\
\cmidrule(lr){2-3}\cmidrule(lr){4-5}
& Score$\uparrow$ & $\Delta$$\uparrow$ & Score$\uparrow$ & $\Delta$$\uparrow$ \\
\midrule
7B Base \cite{liu2023improved} & 1511.47 & -- & 64.60 & -- \\
\ \ +DPO & 1443.20 & -68.27 & 63.14 & -1.46 \\
\rowcolor[gray]{0.92}
\ \ +VAG-DPO & 1459.95 & \textbf{-51.52} & 63.23 & \textbf{-1.37} \\
\midrule
13B Base \cite{liu2023improved} & 1528.02 & -- & 68.47 & -- \\
\ \ +DPO & 1465.29 & -62.73 & 66.31 & -2.16 \\
\rowcolor[gray]{0.92}
\ \ +VAG-DPO & 1473.74 & \textbf{-54.28} & 67.62 & \textbf{-0.85} \\
\bottomrule
\end{tabular}
\end{subtable}

\end{table*}

\noindent
\textbf{Qualitative Case Studies:}
We further provide qualitative evidence to illustrate the practicality of our approach. Figure~\ref{fig:case_study_2} visualizes the cross-attention visual anchors produced by our multi-criteria refinement, showing that the anchors consistently concentrate on query-relevant objects and suppress spurious attention on backgrounds or border regions across diverse scenes. Figure~\ref{fig:qualitative_examples} further compares model outputs and text-to-image cross-attention before and after intervention: when cross-attention becomes unfocused or spatially biased, the baseline tends to introduce unsupported details, whereas Attention-Supervised Inference uses the anchor to re-steer attention toward relevant evidence during later decoding, resulting in more faithful, question-dependent generation. In the car and laptop examples, this correction prevents hallucinated entities/attributes by redirecting attention back to the visually supported regions.

\section{Conclusion}

In this paper, we study hallucinations in MLLMs through the lens of text-to-image cross-attention and identify a consistent signature: cross-attention progressively degrades during generation, shifting decoding toward language priors. 
Based on this analysis, we propose {ADAPT}, which refines multi-layer cross-attention into a query-relevant anchor and leverages it for anchor-guided attention supervision and preference tuning. 
Extensive experiments across architectures and benchmarks show that ADAPT effectively reduces hallucinations with a favorable efficiency--performance trade-off, highlighting cross-attention interventions as a practical path toward more faithful MLLMs.

\section*{Acknowledgements}
This work was supported by the Artificial Intelligence-National Science and Technology Major
Project (2023ZD0121200) and the Fundamental and Interdisciplinary Disciplines Breakthrough Plan of
the Ministry of Education of China (No. JYB2025XDXM103).
%
%
\bibliographystyle{splncs04}
\bibliography{main}

@String(ECCV  = {Eur. Conf. Comput. Vis.})

@String(ICLR  = {Int. Conf. Learn. Represent.})

@article{bai2025qwen2,
  title={{Qwen2.5-VL} Technical Report},
  author={Bai, Shuai and Chen, Keqin and Liu, Xuejing and Wang, Jialin and Ge, Wenbin and Song, Sibo and Dang, Kai and Wang, Peng and Wang, Shijie and Tang, Jun and others},
  journal={arXiv preprint arXiv:2502.13923},
  year={2025}
}

@article{christiano2017deep,
  title={Deep reinforcement learning from human preferences},
  author={Christiano, Paul F and Leike, Jan and Brown, Tom and Martic, Miljan and Legg, Shane and Amodei, Dario},
  journal={Advances in neural information processing systems},
  volume={30},
  year={2017}
}

@article{liu2023llava,
  title={Visual instruction tuning},
  author={Liu, Haotian and Li, Chunyuan and Wu, Qingyang and Lee, Yong Jae},
  journal={Advances in neural information processing systems},
  volume={36},
  pages={34892--34916},
  year={2023}
}

@inproceedings{radford2021learning,
  title={Learning transferable visual models from natural language supervision},
  author={Radford, Alec and Kim, Jong Wook and Hallacy, Chris and Ramesh, Aditya and Goh, Gabriel and Agarwal, Sandhini and Sastry, Girish and Askell, Amanda and Mishkin, Pamela and Clark, Jack and others},
  booktitle={International conference on machine learning},
  pages={8748--8763},
  year={2021},
  organization={PmLR}
}

@article{rafailov2023direct,
  title={Direct preference optimization: Your language model is secretly a reward model},
  author={Rafailov, Rafael and Sharma, Archit and Mitchell, Eric and Manning, Christopher D and Ermon, Stefano and Finn, Chelsea},
  journal={Advances in neural information processing systems},
  volume={36},
  pages={53728--53741},
  year={2023}
}

@inproceedings{rohrbach2018object,
  title={Object hallucination in image captioning},
  author={Rohrbach, Anna and Hendricks, Lisa Anne and Burns, Kaylee and Darrell, Trevor and Saenko, Kate},
  booktitle={Proceedings of the 2018 Conference on Empirical Methods in Natural Language Processing},
  pages={4035--4045},
  year={2018}
}

@inproceedings{sun2023aligning,
  title={Aligning large multimodal models with factually augmented rlhf},
  author={Sun, Zhiqing and Shen, Sheng and Cao, Shengcao and Liu, Haotian and Li, Chunyuan and Shen, Yikang and Gan, Chuang and Gui, Liangyan and Wang, Yu-Xiong and Yang, Yiming and others},
  booktitle={Findings of the Association for Computational Linguistics: ACL 2024},
  pages={13088--13110},
  year={2024}
}

@article{wang2023amber,
  title={{AMBER:} An llm-free multi-dimensional benchmark for mllms hallucination evaluation},
  author={Wang, Junyang and Wang, Yuhang and Xu, Guohai and Zhang, Jing and Gu, Yukai and Jia, Haitao and Wang, Jiaqi and Xu, Haiyang and Yan, Ming and Zhang, Ji and others},
  journal={arXiv preprint arXiv:2311.07397},
  year={2023}
}

@inproceedings{li2023evaluating,
  title={Evaluating object hallucination in large vision-language models},
  author={Li, Yifan and Du, Yifan and Zhou, Kun and Wang, Jinpeng and Zhao, Wayne Xin and Wen, Ji-Rong},
  booktitle={Proceedings of the 2023 conference on empirical methods in natural language processing},
  pages={292--305},
  year={2023}
}

@inproceedings{huang2024opera,
  title={Opera: Alleviating hallucination in multi-modal large language models via over-trust penalty and retrospection-allocation},
  author={Huang, Qidong and Dong, Xiaoyi and Zhang, Pan and Wang, Bin and He, Conghui and Wang, Jiaqi and Lin, Dahua and Zhang, Weiming and Yu, Nenghai},
  booktitle={Proceedings of the IEEE/CVF Conference on Computer Vision and Pattern Recognition},
  pages={13418--13427},
  year={2024}
}

@inproceedings{leng2024mitigating,
  title={Mitigating object hallucinations in large vision-language models through visual contrastive decoding},
  author={Leng, Sicong and Zhang, Hang and Chen, Guanzheng and Li, Xin and Lu, Shijian and Miao, Chunyan and Bing, Lidong},
  booktitle={Proceedings of the IEEE/CVF Conference on Computer Vision and Pattern Recognition},
  pages={13872--13882},
  year={2024}
}

@inproceedings{liu2023improved,
  title={Improved baselines with visual instruction tuning},
  author={Liu, Haotian and Li, Chunyuan and Li, Yuheng and Lee, Yong Jae},
  booktitle={Proceedings of the IEEE/CVF conference on computer vision and pattern recognition},
  pages={26296--26306},
  year={2024}
}

@article{cho2025you,
  title={Do you keep an eye on what i ask? mitigating multimodal hallucination via attention-guided ensemble decoding},
  author={Cho, Yeongjae and Kim, Keonwoo and Hwang, Taebaek and Cho, Sungzoon},
  journal={arXiv preprint arXiv:2505.17529},
  year={2025}
}

@inproceedings{wang2024mdpo,
  title={mdpo: Conditional preference optimization for multimodal large language models},
  author={Wang, Fei and Zhou, Wenxuan and Huang, James Y and Xu, Nan and Zhang, Sheng and Poon, Hoifung and Chen, Muhao},
  booktitle={Proceedings of the 2024 Conference on Empirical Methods in Natural Language Processing},
  pages={8078--8088},
  year={2024}
}

@inproceedings{yu2024rlhf,
  title={Rlhf-v: Towards trustworthy mllms via behavior alignment from fine-grained correctional human feedback},
  author={Yu, Tianyu and Yao, Yuan and Zhang, Haoye and He, Taiwen and Han, Yifeng and Cui, Ganqu and Hu, Jinyi and Liu, Zhiyuan and Zheng, Hai-Tao and Sun, Maosong and others},
  booktitle={Proceedings of the IEEE/CVF Conference on Computer Vision and Pattern Recognition},
  pages={13807--13816},
  year={2024}
}

@article{zhao2023beyond,
  title={Beyond hallucinations: Enhancing lvlms through hallucination-aware direct preference optimization},
  author={Zhao, Zhiyuan and Wang, Bin and Ouyang, Linke and Dong, Xiaoyi and Wang, Jiaqi and He, Conghui},
  journal={arXiv preprint arXiv:2311.16839},
  year={2023}
}

@inproceedings{xie2024v,
  title={V-dpo: Mitigating hallucination in large vision language models via vision-guided direct preference optimization},
  author={Xie, Yuxi and Li, Guanzhen and Xu, Xiao and Kan, Min-Yen},
  booktitle={Findings of the Association for Computational Linguistics: EMNLP 2024},
  pages={13258--13273},
  year={2024}
}

@article{yu2024rlaifv,
  title={Rlaif-v: Aligning mllms through open-source ai feedback for super gpt-4v trustworthiness},
  author={Yu, Tianyu and Zhang, Haoye and Yao, Yuan and Dang, Yunkai and Chen, Da and Lu, Xiaoman and Cui, Ganqu and He, Taiwen and Liu, Zhiyuan and Chua, Tat-Seng and others},
  journal={arXiv preprint arXiv:2405.17220},
  year={2024}
}

@article{liu2024survey,
  title={A survey on hallucination in large vision-language models},
  author={Liu, Hanchao and Xue, Wenyuan and Chen, Yifei and Chen, Dapeng and Zhao, Xiutian and Wang, Ke and Hou, Liping and Li, Rongjun and Peng, Wei},
  journal={arXiv preprint arXiv:2402.00253},
  year={2024}
}

@article{bai2024hallucination,
  title={Hallucination of multimodal large language models: A survey},
  author={Bai, Zechen and Wang, Pichao and Xiao, Tianjun and He, Tong and Han, Zongbo and Zhang, Zheng and Shou, Mike Zheng},
  journal={arXiv preprint arXiv:2404.18930},
  year={2024}
}

@article{zhang2025self,
  title={Self-correcting decoding with generative feedback for mitigating hallucinations in large vision-language models},
  author={Zhang, Ce and Wan, Zifu and Kan, Zhehan and Ma, Martin Q and Stepputtis, Simon and Ramanan, Deva and Salakhutdinov, Russ and Morency, Louis-Philippe and Sycara, Katia and Xie, Yaqi},
  journal={arXiv preprint arXiv:2502.06130},
  year={2025}
}

@article{zou2024look,
  title={Look twice before you answer: Memory-space visual retracing for hallucination mitigation in multimodal large language models},
  author={Zou, Xin and Wang, Yizhou and Yan, Yibo and Lyu, Yuanhuiyi and Zheng, Kening and Huang, Sirui and Chen, Junkai and Jiang, Peijie and Liu, Jia and Tang, Chang and others},
  journal={arXiv preprint arXiv:2410.03577},
  year={2024}
}

@article{li2025mitigating,
  title={Mitigating hallucination for large vision language model by inter-modality correlation calibration decoding},
  author={Li, Jiaming and Zhang, Jiacheng and Jie, Zequn and Ma, Lin and Li, Guanbin},
  journal={arXiv preprint arXiv:2501.01926},
  year={2025}
}

@inproceedings{yu2024attention,
  title={Attention prompting on image for large vision-language models},
  author={Yu, Runpeng and Yu, Weihao and Wang, Xinchao},
  booktitle={European Conference on Computer Vision},
  pages={251--268},
  year={2024},
  organization={Springer}
}

@inproceedings{jiang2025devils,
  title={Devils in middle layers of large vision-language models: Interpreting, detecting and mitigating object hallucinations via attention lens},
  author={Jiang, Zhangqi and Chen, Junkai and Zhu, Beier and Luo, Tingjin and Shen, Yankun and Yang, Xu},
  booktitle={Proceedings of the IEEE/CVF Conference on Computer Vision and Pattern Recognition},
  pages={25004--25014},
  year={2025}
}

@article{kang2025see,
  title={See what you are told: Visual attention sink in large multimodal models},
  author={Kang, Seil and Kim, Jinyeong and Kim, Junhyeok and Hwang, Seong Jae},
  journal={arXiv preprint arXiv:2503.03321},
  year={2025}
}

@article{xing2024mitigating,
  title={Mitigating object hallucination via concentric causal attention},
  author={Xing, Yun and Li, Yiheng and Laptev, Ivan and Lu, Shijian},
  journal={Advances in neural information processing systems},
  volume={37},
  pages={92012--92035},
  year={2024}
}

@article{lyu2024alleviating,
  title={Alleviating hallucinations in large vision-language models through hallucination-induced optimization},
  author={Lyu, Xinyu and Chen, Beitao and Gao, Lianli and Shen, Hengtao and Song, Jingkuan},
  journal={Advances in Neural Information Processing Systems},
  volume={37},
  pages={122811--122832},
  year={2024}
}

@article{huo2024self,
  title={Self-introspective decoding: Alleviating hallucinations for large vision-language models},
  author={Huo, Fushuo and Xu, Wenchao and Zhang, Zhong and Wang, Haozhao and Chen, Zhicheng and Zhao, Peilin},
  journal={arXiv preprint arXiv:2408.02032},
  year={2024}
}

@inproceedings{sarkar2025mitigating,
  title={Mitigating hallucinations in vision-language models through image-guided head suppression},
  author={Sarkar, Sreetama and Che, Yue and Gavin, Alex and Beerel, Peter Anthony and Kundu, Souvik},
  booktitle={Proceedings of the 2025 Conference on Empirical Methods in Natural Language Processing},
  pages={12481--12500},
  year={2025}
}

@inproceedings{zhu2025ibd,
  title={Ibd: Alleviating hallucinations in large vision-language models via image-biased decoding},
  author={Zhu, Lanyun and Ji, Deyi and Chen, Tianrun and Xu, Peng and Ye, Jieping and Liu, Jun},
  booktitle={Proceedings of the Computer Vision and Pattern Recognition Conference},
  pages={1624--1633},
  year={2025}
}

@inproceedings{tang2025seeing,
  title={Seeing far and clearly: Mitigating hallucinations in mllms with attention causal decoding},
  author={Tang, Feilong and Liu, Chengzhi and Xu, Zhongxing and Hu, Ming and Huang, Zile and Xue, Haochen and Chen, Ziyang and Peng, Zelin and Yang, Zhiwei and Zhou, Sijin and others},
  booktitle={Proceedings of the Computer Vision and Pattern Recognition Conference},
  pages={26147--26159},
  year={2025}
}

@inproceedings{woo2025don,
  title={Don't miss the forest for the trees: Attentional vision calibration for large vision language models},
  author={Woo, Sangmin and Kim, Donguk and Jang, Jaehyuk and Choi, Yubin and Kim, Changick},
  booktitle={Findings of the Association for Computational Linguistics: ACL 2025},
  pages={1927--1951},
  year={2025}
}

@inproceedings{wang2025enhancing,
  title={Enhancing visual-language modality alignment in large vision language models via self-improvement},
  author={Wang, Xiyao and Chen, Jiuhai and Wang, Zhaoyang and Zhou, Yuhang and Zhou, Yiyang and Yao, Huaxiu and Zhou, Tianyi and Goldstein, Tom and Bhatia, Parminder and Kass-Hout, Taha and others},
  booktitle={Findings of the Association for Computational Linguistics: NAACL 2025},
  pages={268--282},
  year={2025}
}

@inproceedings{ouali2024clip,
  title={Clip-dpo: Vision-language models as a source of preference for fixing hallucinations in lvlms},
  author={Ouali, Yassine and Bulat, Adrian and Martinez, Brais and Tzimiropoulos, Georgios},
  booktitle={European Conference on Computer Vision},
  pages={395--413},
  year={2024},
  organization={Springer}
}

@inproceedings{subramanian2022reclip,
  title={Reclip: A strong zero-shot baseline for referring expression comprehension},
  author={Subramanian, Sanjay and Merrill, William and Darrell, Trevor and Gardner, Matt and Singh, Sameer and Rohrbach, Anna},
  booktitle={Proceedings of the 60th Annual Meeting of the Association for Computational Linguistics (Volume 1: Long Papers)},
  pages={5198--5215},
  year={2022}
}

@inproceedings{fuyuhan2025mitigatinghall,
  title={Mitigating hallucination in multimodal large language model via hallucination-targeted direct preference optimization},
  author={Fu, Yuhan and Xie, Ruobing and Sun, Xingwu and Kang, Zhanhui and Li, Xirong},
  booktitle={Findings of the Association for Computational Linguistics: ACL 2025},
  pages={16563--16577},
  year={2025}
}

@article{zhao2025tellmodellookmitigating,
  title={Tell Model Where to Look: Mitigating Hallucinations in MLLMs by Vision-Guided Attention},
  author={Zhao, Jianfei and Zhang, Feng and Sun, Xin and Feng, Chong and Tan, Zhixing},
  journal={arXiv preprint arXiv:2511.20032},
  year={2025}
}

@inproceedings{wang2025mllm_can_see,
  author       = {Chenxi Wang and
                  Xiang Chen and
                  Ningyu Zhang and
                  Bozhong Tian and
                  Haoming Xu and
                  Shumin Deng and
                  Huajun Chen},
  title        = {{MLLM} can see? Dynamic Correction Decoding for Hallucination Mitigation},
  booktitle    = {The Thirteenth International Conference on Learning Representations,
                  {ICLR} 2025, Singapore, April 24-28, 2025},
  publisher    = {OpenReview.net},
  year         = {2025},
  bibsource    = {dblp computer science bibliography, https://dblp.org}
}

@article{tong2025layercd,
  title={Mitigating Hallucination in Multimodal LLMs with Layer Contrastive Decoding},
  author={Tong, Bingkui and Xia, Jiaer and Zhou, Kaiyang},
  journal={arXiv preprint arXiv:2509.25177},
  year={2025}
}

@inproceedings{sarkar2025dpa,
  author       = {Pritam Sarkar and
                  Sayna Ebrahimi and
                  Ali Etemad and
                  Ahmad Beirami and
                  Sercan {\"{O}}. Arik and
                  Tomas Pfister},
  title        = {Mitigating Object Hallucination in MLLMs via Data-augmented Phrase-level
                  Alignment},
  booktitle    = {The Thirteenth International Conference on Learning Representations,
                  {ICLR} 2025, Singapore, April 24-28, 2025},
  year         = {2025},
  bibsource    = {dblp computer science bibliography, https://dblp.org}
}

@article{singh2025openaigpt5card,
  title={{OpenAI} {GPT-5} System Card},
  author={Singh, Aaditya and Fry, Adam and Perelman, Adam and Tart, Adam and Ganesh, Adi and El-Kishky, Ahmed and McLaughlin, Aidan and Low, Aiden and Ostrow, AJ and Ananthram, Akhila and others},
  journal={arXiv preprint arXiv:2601.03267},
  year={2025}
}

@inproceedings{fu2025mmecomprehensiveevaluationbenchmark,
  author       = {Chaoyou Fu and
                  Peixian Chen and
                  Yunhang Shen and
                  Yulei Qin and
                  Mengdan Zhang and
                  Xu Lin and
                  Jinrui Yang and
                  Xiawu Zheng and
                  Ke Li and
                  Xing Sun and
                  Yunsheng Wu and
                  Rongrong Ji and
                  Caifeng Shan and
                  Ran He},
  editor       = {Danielle Belgrave and
                  Cheng Zhang and
                  Laura N. Montoya and
                  Hsuan{-}Tien Lin and
                  Razvan Pascanu and
                  Piotr Koniusz and
                  Marzyeh Ghassemi and
                  Nancy Chen and
                  Iv{\'{a}}n Vladimir Meza Ru{\'{\i}}z and
                  Arturo Loaiza{-}Bonilla},
  title        = {{MME:} {A} Comprehensive Evaluation Benchmark for Multimodal Large
                  Language Models},
  booktitle    = {Advances in Neural Information Processing Systems},
  year         = {2025},
  bibsource    = {dblp computer science bibliography, https://dblp.org}
}

@inproceedings{liu2024mmbenchmultimodalmodelallaround,
  author       = {Yuan Liu and
                  Haodong Duan and
                  Yuanhan Zhang and
                  Bo Li and
                  Songyang Zhang and
                  Wangbo Zhao and
                  Yike Yuan and
                  Jiaqi Wang and
                  Conghui He and
                  Ziwei Liu and
                  Kai Chen and
                  Dahua Lin},
  title        = {MMBench: Is Your Multi-modal Model an All-Around Player?},
  booktitle    = {{ECCV} {(6)}},
  series       = {Lecture Notes in Computer Science},
  volume       = {15064},
  pages        = {216--233},
  publisher    = {Springer},
  year         = {2024}
}
\end{document}